\documentclass[lettersize,journal]{IEEEtran}
\usepackage{amsmath,amsfonts}
\usepackage{algorithmic}
\usepackage{algorithm}
\usepackage{array}
\usepackage[caption=false,font=normalsize,labelfont=sf,textfont=sf]{subfig}
\usepackage{textcomp}
\usepackage{stfloats}
\usepackage{url}
\usepackage{verbatim}
\usepackage{graphicx}
\usepackage{cite}
\usepackage{booktabs}
\usepackage{threeparttable}
\usepackage{multirow}

\usepackage{cases}
\usepackage{bbm}
\usepackage{amssymb}

\usepackage{color}

\begin{document}


\title{IVAC-$\mathbf{P^2L}$:~Leveraging Irregular Repetition Priors for Improving Video Action Counting}

\author{Hang~Wang,
        Zhi-Qi~Cheng,~\IEEEmembership{Member,~IEEE,}
        Youtian~Du,~\IEEEmembership{Member,~IEEE,}
        and~Lei~Zhang,~\IEEEmembership{Fellow,~IEEE}
\thanks{This work was completed through collaboration with Carnegie Mellon University's Language Technologies Institute. (Corresponding authors: Zhi-Qi Cheng; Youtian Du.)}
\thanks{H. Wang is with Xi’an Jiaotong University, China, and The Hong Kong Polytechnic University, Hong Kong (e-mail: cshangwang@xjtu.edu.cn).}
\thanks{Z.-Q. Cheng is with the Language Technologies Institute, Carnegie Mellon University, Pittsburgh, PA, USA (e-mail: zhiqic@cs.cmu.edu).}
\thanks{Y. Du is with Xi’an Jiaotong University, China (e-mail: duyt@mail.xjtu.edu.cn).}
\thanks{L. Zhang is with The Hong Kong Polytechnic University, Hong Kong (e-mail: cslzhang@comp.polyu.edu.hk).}
}

\markboth{Journal of \LaTeX\ Class Files,~Vol.~14, No.~8, August~2021}%
{Shell \MakeLowercase{\textit{et al.}}: A Sample Article Using IEEEtran.cls for IEEE Journals}


\maketitle

\begin{abstract}
The quantification of repetitive actions within videos, a task commonly referred to as Video Action Counting (VAC), is a critical challenge in understanding and analyzing content within sports, fitness, and daily activities. Traditional approaches to VAC have largely overlooked the nuanced irregularities inherent in action repetitions, such as interruptions and variable lengths between cycles. Addressing this gap, our study introduces a novel perspective on VAC, focusing on Irregular Video Action Counting (IVAC), which emphasizes the importance of modeling the irregular repetition priors present in video content.
We conceptualize these priors through two key aspects: \textit{Inter-cycle Consistency} and \textit{Cycle-interval Inconsistency}. Inter-cycle Consistency ensures that spatial-temporal representations of all cycle segments are homogeneous, reflecting the uniformity of actions within cycles. In contrast, Cycle-interval Inconsistency mandates a clear semantic distinction between the representations of cycle segments and intervals, acknowledging the inherent dissimilarities in content. To effectively encapsulate these priors, we introduce a novel methodology comprising consistency and inconsistency modules, underpinned by a tailored pull-push loss ($\mathbf{P^2L}$) mechanism. This approach employs a pull loss to enhance the cohesion among cycle segment features and a push loss to differentiate between cycle and interval segment features distinctly.
Empirical evaluations on the RepCount dataset illustrate that our IVAC-$\mathbf{P^2L}$ model sets a new benchmark in state-of-the-art performance for the VAC task. Moreover, our model demonstrates exceptional adaptability and generalization across diverse video content, achieving superior performance on two additional datasets, UCFRep and Countix, without necessitating dataset-specific fine-tuning. These findings not only validate the effectiveness of our approach in addressing the complexities of irregular repetition in videos but also open new avenues for future research in video understanding and analysis.\footnote{Source code:~\url{https://github.com/hwang-cs-ime/IVAC-P2L}}
\end{abstract}


\section{Introduction}
\label{introduction}
\IEEEPARstart{R}{epetitive} actions are a fundamental aspect of both natural phenomena and human activities. From the grand cosmic dance of planetary rotations governed by Newton's laws of motion to the life-sustaining rhythm of heartbeats detectable by modern sensors, repetition is ubiquitous. In the realm of human endeavors, repetitive motions span a wide spectrum, including culinary tasks like slicing onions, athletic activities such as trampolining, and intricate assembly processes in manufacturing. Unlike the predictable cycles of celestial bodies or the measurable beats of a heart, the counting of actions in manufacturing and other human-centric activities often relies on the analysis of visual data through video surveillance. This necessitates advancements in Video Action Counting (VAC), a computer vision approach aimed at discerning the frequency of repetitive actions within video footage. As a non-invasive method that leverages readily available camera data, VAC presents a versatile tool for understanding complex activities without the need for physical or intrusive sensors \cite{soro2019recognition}. Beyond its direct applications, the insights gained from VAC contribute significantly to various domains of video analysis, including event detection \cite{vats2020event, Kang_2022_CVPR, 9880248, 9880059}, pedestrian detection \cite{ran2007pedestrian, Zhang_2021_CVPR, Lima_2021_CVPR}, and 3D reconstruction \cite{ribnick20103d, ribnick2012reconstructing, wandt20163d, li2018structure}. Despite its potential, VAC is challenged by two critical aspects: the \emph{Class-agnostic Characteristic}, requiring the model to accurately count actions regardless of their specific nature; and \emph{Spatial-temporal Irregularity}, reflecting the varied lengths of intervals between actions and inconsistencies in the execution speed and completion of the repetitive actions themselves.

\begin{figure}[t]
\centering
    \includegraphics[width=0.98\columnwidth]{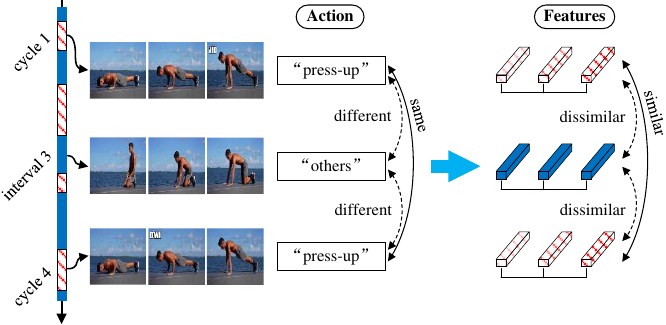} 
\caption{Conceptual illustration of the IVAC-$\mathtt{P^2L}$ approach. This figure highlights the core principle that underpins our model: the inherent similarity in spatial-temporal features among cycle segments due to their shared action, contrasted with the fundamental dissimilarity between the features of cycle and interval segments, reflecting the distinct nature of the actions they encapsulate. This duality forms the basis for our pull-push loss mechanism, aimed at accurately distinguishing and counting repetitive actions amidst variability and interruptions.}
\label{fig1}
\end{figure}

Despite the advances in video action counting, much of the existing literature \cite{OfirLevy2015LiveRC, AlexiaBriassouli2007ExtractionAA, RossCutler2000RobustRP, XiaoxiaoLi2018RepetitiveME} has primarily addressed challenges related to class-agnostic characteristics, often transforming VAC into problems of classification, regression, and detection. Levy \& Wolf \cite{OfirLevy2015LiveRC} innovatively used multiple detectors for cycle segment identification, a method that struggles with the complexity of variable-length cycles. Zhang et al. \cite{zhang2020context} enhanced the adaptability to varied cycle lengths by leveraging contextual cues, a significant step toward more nuanced action understanding. Similarly, Dwibedi et al. \cite{dwibedi2020counting} utilized a self-similarity matrix to analyze frame periodicity and cycle lengths, contributing a valuable perspective on temporal pattern recognition. Addressing spatial-temporal irregularities directly, Hu et al. \cite{hu2022transrac} introduced the RepCount dataset, enriched with detailed annotations, and proposed a density map-based counting method, marking a pioneering effort to capture the intricacies of action repetitions in more lifelike settings. However, this focus on the framework did not fully exploit the potential of modeling the underlying irregular repetition priors present in video sequences, particularly the pronounced spatial-temporal consistency across different cycle segments indicative of repetitive actions. Our work seeks to bridge this gap by emphasizing the critical need to model both the uniformity within cycle segments and the variance between cycles and intervals, laying the foundation for a more comprehensive approach to VAC.

In response to the identified gaps in the current video action counting (VAC) methodologies, this study introduces a groundbreaking approach, IVAC-$\mathtt{P^2L}$, designed to meticulously model the spatial-temporal irregularities intrinsic to repetitive actions in videos. Central to our method is the distinction between cycle and interval segments, rooted in the premise that repetitions of the same action across different cycles yield highly similar spatial-temporal features, despite the potential disruptions posed by intervening non-repetitive segments (Fig. \ref{fig1}). This observation has led us to define two fundamental irregular repetition priors that underlie the VAC task: \emph{Inter-cycle Consistency} and \emph{Cycle-interval Inconsistency}. Inter-cycle Consistency posits that cycle segments, owing to their shared action, should exhibit close spatial-temporal feature alignment, whereas Cycle-interval Inconsistency asserts that the distinctive actions captured in interval segments necessitate a clear demarcation in feature space from those of cycle segments.

To operationalize these priors, we innovatively employ mean-pooling to aggregate the visual features of all frames within cycle and interval segments, thereby extracting their quintessential spatial-temporal characteristics. Building upon this foundation, we introduce dedicated consistency and inconsistency modules aimed at encapsulating the nuanced dynamics of irregular repetitions. At the heart of our model lies the novel pull-push loss mechanism ($\mathtt{P^2L}$), comprising a pull loss that synergizes the features across cycle segments to foster mutual closeness, and a push loss that strategically distances the feature representations of cycle segments from those of intervals. This dual loss framework not only enhances the model's ability to discern repetitive from non-repetitive segments but also significantly elevates the precision of action counting in videos, marking a substantial advancement in the field.

To encapsulate, the contributions of our work are multifaceted and significant, encompassing the development of a novel framework and its empirical validation. Specifically, our contributions can be summarized as follows:
\begin{itemize}
    \item We introduce a conceptualization of irregular repetition priors within the domain of video action counting, underpinned by our IVAC-$\mathtt{P^2L}$ model. This model adeptly captures the nuanced dynamics of repetitive actions through two distinct yet complementary facets: \textit{Inter-cycle Consistency} and \textit{Cycle-interval Inconsistency}. These facets collectively facilitate a more refined and accurate representation of action repetitions, setting a new precedent for subsequent research.

    \item To operationalize these concepts, we architect specialized consistency and inconsistency modules within the IVAC-$\mathtt{P^2L}$ framework, augmented by a bespoke pull-push loss mechanism. This loss function coordinates the spatial-temporal features across cycle segments while delineating them from interval segments, effectively encapsulating the essence of irregular repetition. Our ablation studies underscore the efficacy of this approach, demonstrating its superiority over existing loss mechanisms in capturing the complexities of video action counting.

    \item Our comprehensive empirical evaluation showcases the exceptional performance of the IVAC-$\mathtt{P^2L}$ model across diverse datasets. Notably, it achieves unparalleled accuracy on the RepCount dataset and exhibits robust generalizability to the UCFRep and Countix datasets without necessitating dataset-specific fine-tuning. These results not only verify the effectiveness of our approach but also highlight its potential applicability across a wide array of video analysis tasks.
\end{itemize}
In sum, the IVAC-$\mathtt{P^2L}$ framework represents an advancement in video action counting, heralding a new era of research that rigorously addresses the intricacies of irregular repetitions in video content.

\begin{figure*}[t]
\centering
    \includegraphics[width=0.95\textwidth]{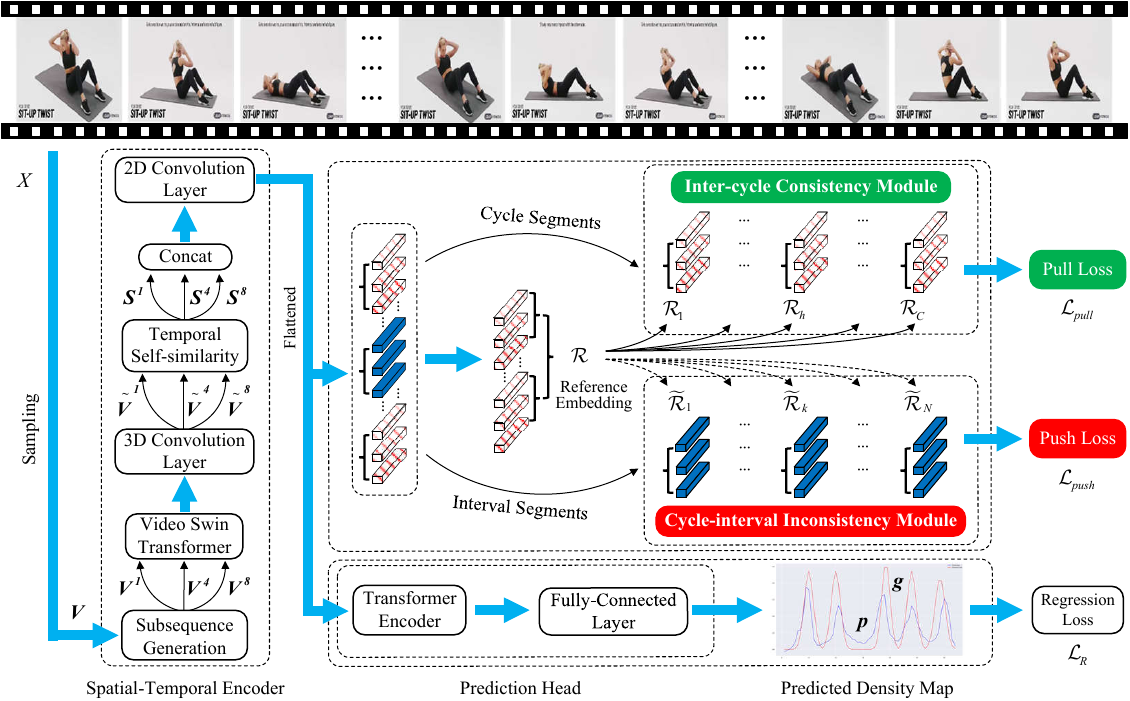}
\caption{Architectural overview of IVAC-$\mathtt{P^2L}$. This diagram delineates the integrated structure of IVAC-$\mathtt{P^2L}$, showcasing its principal components: the spatial-temporal encoder, prediction head, inter-cycle consistency module, and cycle-interval inconsistency module. Initially, the spatial-temporal encoder extracts nuanced features from the video, which the prediction head processes to generate a density map, facilitating accurate action counting. The inter-cycle consistency module ensures homogeneity among features of cycle segments, reflecting their repetitive nature, while the cycle-interval inconsistency module distinguishes these cycle segments from non-repetitive interval segments, leveraging the pull-push loss mechanism to enhance counting precision and reliability.}
\label{fig2}
\end{figure*}

\section{Related Work}
This section outlines the previous works and recent advancements in video activity analysis, specifically focusing on video action recognition, detection, temporal localization, spatial-temporal grounding, and directly related to our work, video action counting. Additionally, we explore the application of contrastive learning in video representation learning.

\subsection{Video Activity Analysis}
Video activity analysis stands as a comprehensive field aimed at deciphering the complex dynamics of objects and humans within videos across both temporal and spatial dimensions. This domain encompasses several pivotal research areas, each contributing uniquely towards understanding video content. Key areas include video action recognition \cite{wu2022memvit, Wang_2022_CVPR, 9381645}, which focuses on assigning action labels to videos by interpreting the dynamic interplay between subjects and their environment. Video action detection \cite{chen2021watch, zhao2022tuber, 9754228} extends this challenge by identifying the temporal boundaries and categories of actions within untrimmed footage, presenting a higher level of complexity. Another closely related area, temporal video localization \cite{Nan_2021_CVPR, 9316901, 10058582, 10233190}, seeks to pinpoint the start and end of actions based on descriptive expressions, while spatial-temporal video grounding \cite{9446308, Su_2021_ICCV, yang2022tubedetr} further complicates this by requiring the localization of actions and objects as specified by textual queries.

Distinct from these tasks, video action counting emerges as a class-agnostic challenge that focuses on quantifying repetitive actions within a video. This task necessitates the identification and enumeration of actions that, despite belonging to the same category, may vary significantly in amplitude and frequency. Unlike action recognition and detection, video action counting does not classify the actions but rather assesses their occurrence, presenting a unique blend of complexities. Moreover, video action counting's emphasis on repetition counting without specific content classification sets it apart from temporal localization and spatial-temporal grounding, highlighting its unique position within the broader spectrum of video activity analysis. This specialized focus on repetition quantification underscores the nuanced differences between video action counting and other video analysis tasks, illustrating the necessity for targeted approaches in addressing its challenges.

\subsection{Video Action Counting}
Recent years have witnessed a surge in interest within the computer vision community towards the nuanced task of video action counting. This task, aimed at quantifying repetitive actions within videos, has seen a variety of innovative approaches aimed at addressing its inherent challenges. Initial strategies, such as the online framework by Levy \& Wolf \cite{levy2015live}, focused on the automatic detection of cycle segments' start and end points through a shifting window mechanism. Runia et al. \cite{runia2018real} advanced this domain by tailoring their approach to non-static and non-stationary video conditions, applying continuous wavelet transforms to optical flow features for more accurate action repetition estimation.
The complexity of video action counting was further explored through the work of Yin et al. \cite{yin2021energy}, who leveraged pre-trained action recognition models in conjunction with PCA for generating periodic signals indicative of repetitive motion. Dwibedi et al. \cite{dwibedi2020counting} introduced a novel perspective by employing a temporal self-similarity matrix, marking a significant step towards understanding cycle length and periodicity within video sequences. To accommodate the variance in cycle lengths found in real-world scenarios, Zhang et al. \cite{zhang2020context} devised a context-aware model that utilized the estimated positions of adjacent cycles to refine action counting accuracy.

Further bridging the gap between academic research and practical applications, Hu et al. \cite{hu2022transrac} presented the RepCount dataset, enriched with fine-grained annotations that capture long-range videos and the interruptions found within inconsistent cycle segments. This development underscored the necessity for methodologies capable of navigating the complex dynamics of real-life video content. Additionally, Jacquelin et al. \cite{jacquelin2022periodicity} explored unsupervised methods for repetition counting, extending the applicability of these techniques to time-series data and showcasing the potential for broader generalization.
Recognizing the limitations of visual data under challenging conditions, subsequent studies have begun to incorporate auxiliary signals such as audio and human poses to enhance the robustness of video action counting. Ferreira et al. \cite{ferreira2021deep} were pioneers in integrating human pose estimation results, while Yao et al. \cite{yao2023poserac} emphasized the role of salient poses in representing actions for more effective repetition analysis. Zhang et al. \cite{zhang2021repetitive} further expanded the modality palette by fusing audio and visual signals, demonstrating the benefits of a multimodal approach in overcoming the shortcomings of visual data alone.

Our work continues this trajectory by concentrating on the unique challenges presented by videos featuring irregular repetitions. We aim to harness irregular repetition priors to enhance video action counting performance, adopting a cross-modal semantic perspective that prioritizes the nuanced understanding of action dynamics over simple quantitative assessment. By doing so, we aspire to advance the field of video action counting towards greater accuracy and applicability in real-world scenarios.

\subsection{Contrastive Learning}
The principle of contrastive learning, which seeks to structure an embedding space where similar samples converge while dissimilar ones diverge, has markedly influenced the realms of both image and video analysis \cite{dai2017contrastive, zhang2019graphical, park2020contrastive, kamath2021mdetr, pan2021videomoco, qian2021spatiotemporal, zeng2021contrastive}. This learning paradigm has demonstrated exceptional efficacy across various tasks, significantly advancing the capabilities in distinguishing nuanced differences and similarities among data samples. In the field of image analysis, Zhang et al. \cite{zhang2019graphical} innovated by forming contrastive pairs to tackle the challenges of instance confusion and relationship ambiguity in scene graph generation. This method underscores the potential of contrastive learning to enhance semantic understanding and refine relational mappings within complex visual contexts. Similarly, Kamath et al. \cite{kamath2021mdetr} leveraged contrastive alignments between object bounding boxes and textual tokens to enrich the comprehension of referring expressions, illustrating the versatility of contrastive learning in bridging visual and linguistic domains.
Transitioning to video representation learning, the application of contrastive learning extends to generating insightful spatial-temporal embeddings through techniques like data augmentation \cite{qian2021spatiotemporal, pan2021videomoco} and extraction from temporally distinct blocks \cite{zeng2021contrastive}. These methods have proven instrumental in cultivating a more profound understanding of video content by emphasizing temporal dynamics and enhancing the differentiation of actions over time.

Inspired by these advancements, our approach in the video action counting domain adopts the foundational principles of contrastive learning. We classify cycle and interval segments within videos as analogous to positive and negative samples, respectively. This conceptualization allows us to exploit the semantic disparities inherent between repetitive actions and intervening non-action segments, utilizing the contrastive framework to emphasize these distinctions in spatial-temporal representation. By doing so, we aim to harness the discriminative power of contrastive learning to refine the identification and quantification of repetitive actions, thereby advancing the precision and robustness of video action counting methodologies.

\section{Methodology}
Our exploration of the Video Action Counting (VAC) task has unveiled two pivotal challenges: the \textit{class-agnostic characteristic} and the \textit{spatial-temporal irregularity} of action repetitions. To surmount these hurdles, we devise a comprehensive methodology articulated through three principal phases: (i)~reformulating the VAC problem within an end-to-end regression framework, (ii)~introducing novel modules to model irregular repetition priors, and (iii)~delineating the backbone network architecture in tandem with a novel data augmentation strategy. Fig.~\ref{fig2} presents an illustrative overview of the proposed IVAC-$\mathtt{P^2L}$ framework.

In this section, we first elucidate our problem formulation, redefining VAC as a regression task to directly address the class-agnostic challenge (Section~\ref{subsec:problem_formulation}). Subsequently, we introduce innovative modules that leverage irregular repetition priors to handle spatial-temporal irregularities, underpinned by a novel pull-push loss function (Section~\ref{subsec:irregular_priors}). Furthermore, we detail our spatial-temporal encoder and prediction head, designed to accommodate diverse action frequencies, durations, and intensities (Section~\ref{subsec:encoder_head}). Finally, we propose a novel data augmentation strategy, termed Random Count Augmentation (RCA), to enhance the model's generalization and robustness across various VAC scenarios (Section~\ref{subsec:rca}).

\subsection{Problem Formulation}
\label{subsec:problem_formulation}
To confront the \textit{class-agnostic} challenge, we propose an end-to-end framework that reconceptualizes the Video Action Counting (VAC) task as a regression problem. Given an input video $X$ comprising $m$ frames, denoted as $X = \{x^1, x^2, \cdots, x^m\}$, our objective is to estimate the total number of repetitive actions within $X$. This formulation is expressed as:
\begin{equation}
    T = \Psi(X; \Theta),
    \label{eq:vac_regression}
\end{equation}
where $X$ represents the input video, $\Psi(\cdot; \Theta)$ symbolizes our regression model with learnable parameters $\Theta$, and $T$ is the estimated count of repetitive actions. By formulating VAC as a regression task focused on count estimation rather than action classification, we directly address the class-agnostic nature of the problem, circumventing the need for explicit action labels or categories.

To tackle the \textit{spatial-temporal irregularity} challenge, which arises due to the irregular and diverse patterns of action repetitions across videos, we propose an optimization strategy involving two key objectives: \textit{Inter-cycle Consistency} and \textit{Cycle-Interval Inconsistency}. The \textit{Inter-cycle Consistency} objective ensures that representations of segments within the same action cycle exhibit a high degree of similarity, thereby enforcing a consistent depiction of repetitive actions across the video. This consistency objective is crucial for accurately capturing and counting the repetitive patterns, despite potential variations in action execution or duration.

Conversely, the \textit{Cycle-Interval Inconsistency} objective distinguishes between action cycles (repetitive action segments) and intervals of non-action or unrelated activities, pushing their representations apart in the feature space. This dissimilarity objective enhances the model's discriminative capability, enabling it to differentiate between repetitive actions and irrelevant or non-repetitive segments, which is essential for accurate action counting in the presence of spatial-temporal irregularities.

To optimize our framework, we articulate the overall training objective as:
\begin{equation}
    \mathcal{L} = \underbrace{\mathcal{L}_{Pull} + \mathcal{L}_{Push}}_{{\mathcal{L}}_{P}} + \mathcal{L}_{R},
    \label{eq:overall_loss}
\end{equation}
where $\mathcal{L}_P$ emerges from enforcing constraints on the relationships between cycle (repetitive action) and interval (non-action or unrelated activity) segments, leveraging repetition priors to enhance model performance. The dual-component loss $\mathcal{L}_P$ facilitates the modeling of \textit{inter-cycle consistency} ($\mathcal{L}_{Pull}$) and \textit{cycle-interval inconsistency} ($\mathcal{L}_{Push}$), integral to our strategy for addressing spatial-temporal irregularities. $\mathcal{L}_R$ is a regression loss, common across VAC techniques, aimed at refining the accuracy of predicted action counts.

By combining the regression formulation (Eq. \ref{eq:vac_regression}) with the dual-objective optimization strategy (Eq. \ref{eq:overall_loss}), our framework tackles both the class-agnostic and spatial-temporal irregularity challenges inherent to the VAC task. This formulation allows the model to learn robust representations that capture the essence of repetitive actions while distinguishing them from non-repetitive segments, ultimately enabling accurate action counting across diverse video content.

\subsection{Irregular Repetition Priors and Pull-Push Loss}
\label{subsec:irregular_priors}
Effectively handling the \textit{spatial-temporal irregularity} challenge is crucial for accurate video action counting. This irregularity arises from the diverse patterns and variations in action repetitions across different videos, including inconsistencies in action duration, execution speed, and temporal spacing between repetitions. To tackle this challenge, we propose a novel approach that leverages irregular repetition priors and models the inter-cycle consistency and cycle-interval inconsistency within video sequences.

\subsubsection{Reference Embeddings}
The foundation of our approach lies in accurately distinguishing between repetitive actions (cycle segments) and non-repetitive segments (intervals) within a video $X$. This distinction is critical for precise action counting and comprehending the overall action sequence. To this end, we introduce a nuanced method for generating reference embeddings that encapsulate the representative characteristics of both cycle and interval segments.

For each cycle segment $h$ in $X$, identified by its index, we calculate a reference embedding $\mathcal{R}_h$ that captures the essence of the repetitive action present. This is achieved through mean-pooling the feature vectors across all frames within the segment, as shown in Equation~\eqref{eq3}:
\begin{equation}
    \mathcal{R}_h = \frac{1}{|\mathcal{C}_h|} \sum_{i \in \mathcal{C}_h} \boldsymbol{E}_{i,:},
    \label{eq3}
\end{equation}
where $\boldsymbol{E}_{i,:}$ denotes the features of the $i$-th frame, and $\mathcal{C}_h$ represents the set of frames comprising the $h$-th cycle segment. The resulting embedding $\mathcal{R}_h$ serves as a comprehensive representation of the cycle's spatial-temporal characteristics, capturing the intrinsic patterns and dynamics of the repetitive action.

To synthesize a global perspective of repetitive actions across the entire video, we aggregate the embeddings of all identified cycle segments, yielding a collective reference embedding $\mathcal{R}$, as defined in Equation~\eqref{eq4}:
\begin{equation}
    \mathcal{R} = \frac{1}{C} \sum_{h=1}^{C} \mathcal{R}_h,
    \label{eq4}
\end{equation}
where $C$ is the total number of cycle segments within the video. This collective embedding $\mathcal{R}$ serves as a reference point for assessing the consistency of repetitive actions throughout the video sequence, enabling the model to establish a unified representation of the core action being repeated.

Similarly, for the $k$-th interval segment, which may contain non-action or distinct action intervals, we compute a reference embedding $\widetilde{\mathcal{R}}_k$ by mean-pooling frame features within that interval:
\begin{equation}
    \widetilde{\mathcal{R}}_k = \frac{1}{|\mathcal{N}_k|} \sum_{i \in \mathcal{N}_k} \boldsymbol{E}_{i,:},
    \label{eq5}
\end{equation}
where $\mathcal{N}_k$ encapsulates the frames within the $k$-th interval segment. This embedding $\widetilde{\mathcal{R}}_k$ captures the unique spatial-temporal features of intervals, facilitating the model's ability to differentiate between repetitive and non-repetitive segments within the video sequence.

These reference embeddings play a pivotal role in the model's optimization process, as they enable the enforcement of two critical principles: inter-cycle consistency and cycle-interval inconsistency, as described next.

\begin{figure}[t]
\centering
    \includegraphics[width=0.98\columnwidth]{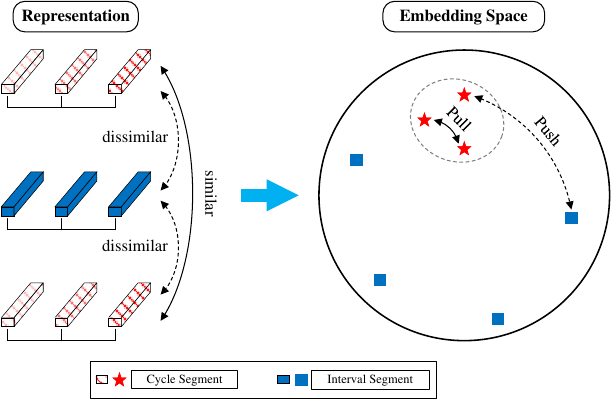} 
\caption{Conceptual visualization of Inter-cycle Consistency and Cycle-interval Inconsistency mechanisms. On the left, we showcase the process of extracting spatial-temporal features from different video segments, illustrating how these features form the basis for our analysis. The right subfigure then translates these extracted features into an embedding space, visually demonstrating the principle of inter-cycle consistency by grouping similar cycle segments closer together and enforcing cycle-interval inconsistency by distancing cycle segments from distinct interval segments. This dual representation underscores the core methodology of our approach, emphasizing the strategic separation and aggregation of features to accurately count and differentiate between repetitive actions and non-repetitive segments.}
\label{fig3}
\end{figure}

\subsubsection{Inter-cycle Consistency and Pull Loss}
The concept of inter-cycle consistency is key in accurately distinguishing and counting repetitive actions within a video. It is predicated on the assumption that all instances of a given action, despite potential minor variations, share a core semantic similarity that can be captured through their spatial-temporal representations. This similarity implies that the feature embeddings of cycle segments, which correspond to repetitions of the same action, should converge in the embedding space, reflecting a high degree of uniformity.

To quantitatively enforce this principle, we introduce the pull loss, $\mathcal{L}_{Pull}$, designed to minimize the distance between the embedding of each cycle segment, $\mathcal{R}_h$, and the collective reference embedding, $\mathcal{R}$, representative of all cycle segments within the video. This loss is formally defined as:
\begin{equation}
    \mathcal{L}_{Pull} = \frac{1}{C} \sum_{h=1}^{C} \left( 1 - \cos(\mathcal{R}_h, \mathcal{R}) \right),
    \label{eq6}
\end{equation}
where $\cos(\cdot,\cdot)$ denotes the cosine similarity between two embeddings, and $C$ is the total number of cycle segments identified in the video. The pull loss aims to maximize the cosine similarity (or minimize the cosine distance) between the embeddings of individual cycle segments and the global cycle reference embedding, thereby encouraging a closer representation of repetitive actions in the feature space.

The operationalization of inter-cycle consistency through the pull loss serves multiple purposes within our model's architecture. Firstly, it ensures that the model recognizes and aligns the representations of repeated actions, facilitating a more accurate and robust counting mechanism. Secondly, by embedding this consistency criterion directly into the loss function, the model is guided to learn representations that inherently capture the essence of repetitive actions, improving its generalizability and performance across diverse video content. Furthermore, this approach allows the model to effectively handle variations within repetitions, accommodating for minor discrepancies in action execution while still maintaining a strong semantic linkage across instances. The end result is a model that not only excels in counting actions with high precision but also exhibits resilience in the face of spatial-temporal irregularities present in real-world videos.

By enforcing inter-cycle consistency through the pull loss, our model is able to learn a unified representation of the core repetitive action, while simultaneously accounting for potential variations in execution. This consistency principle plays a crucial role in accurately identifying and counting repetitive actions, even in the presence of spatial-temporal irregularities that may arise from diverse action patterns, durations, and temporal spacing across different videos.

\subsubsection{Cycle-Interval Inconsistency and Push Loss}

It is crucial to accurately differentiate between repetitive action sequences (cycle segments) and non-action or distinct action intervals (interval segments). This differentiation is based on the principle of cycle-interval inconsistency, which posits that the spatial-temporal characteristics of interval segments should be significantly different from those of cycle segments. This inconsistency ensures that the model can effectively distinguish and accurately count discrete actions without any confusion.

To operationalize this principle within our model, we introduce the push loss, $\mathcal{L}_{Push}$, designed to amplify the dissimilarity between the embeddings of interval segments and the collective reference embedding of cycle segments. The push loss, across $N$ interval segments in the video, is mathematically defined as:
\begin{equation}
    \mathcal{L}_{Push} = \frac{1}{N} \sum_{k=1}^{N} e^{- \left( 1 - \cos(\widetilde{\mathcal{R}}_k, \mathcal{R}) \right)},
    \label{eq7}
\end{equation}
where $\cos(\cdot, \cdot)$ measures the cosine similarity between two embeddings, $\widetilde{\mathcal{R}}_k$ represents the reference embedding of the $k$-th interval segment, and $\mathcal{R}$ is the collective reference embedding of cycle segments. By maximizing the exponential negative cosine similarity, this loss function effectively ``pushes" the embeddings of interval segments away from those of cycle segments in the embedding space, enhancing the model's ability to differentiate between them.

Complementing the inter-cycle consistency objective enforced by the pull loss, the cycle-interval inconsistency principle and corresponding push loss enable our model to accurately distinguish between repetitive actions and non-repetitive segments within a video. This distinction is crucial for precise action counting, as it prevents the model from confusing non-action intervals with repetitive cycles, thereby avoiding erroneous count predictions.

\subsubsection{Regression Loss}
In conjunction with the pull and push losses, we employ a regression loss, $\mathcal{L}_R$, based on mean squared error (MSE), to fine-tune the model's predictions against the ground-truth action counts. This is expressed as:
\begin{equation}
    \mathcal{L}_R = \frac{1}{L} \sum_{i=1}^{L} \left( p_i - g_i \right)^2,
    \label{eq8}
\end{equation}
where $p_i$ corresponds to the predicted action count at the $i$-th frame, $g_i$ denotes the ground-truth count, and $L$ is the total number of frames in video $X$. This loss ensures that the model's predictions are closely aligned with the actual action frequencies within the video, further refining the accuracy of the count estimates.

The comprehensive training objective for our model combines the pull, push, and regression loss components to optimize both the discriminative capability and the action counting accuracy:
\begin{equation}
    \mathcal{L} = \alpha \mathcal{L}_{Pull} + \beta \mathcal{L}_{Push} + \gamma \mathcal{L}_R,
    \label{eq9}
\end{equation}
where $\alpha$, $\beta$, and $\gamma$ are hyperparameters that balance the contributions of the respective loss components. This formulation enables a holistic optimization strategy that addresses both the spatial-temporal irregularities and the semantic distinctions between cycle and interval segments. By implementing the push loss in concert with the pull loss and regression loss, our model achieves a nuanced understanding of action sequences within videos, effectively handling the class-agnostic and spatial-temporal irregularity challenges inherent to the VAC task. Through the synergistic enforcement of inter-cycle consistency, cycle-interval inconsistency, and regression objectives, our model learns robust representations that capture the essence of repetitive actions while simultaneously distinguishing them from non-repetitive segments. This integrated approach allows for accurate action counting across diverse video content, exhibiting resilience to the spatial-temporal irregularities prevalent in real-world scenarios.

\subsection{Spatial-Temporal Encoder and Prediction Head}
\label{subsec:encoder_head}
In addressing the diversity of action frequencies, durations, and intensities across various videos, our approach employs a multi-scale subsequence generation technique, feeding these subsequences into a spatial-temporal encoder for comprehensive feature extraction. This design allows our model to accommodate and effectively capture the rich variations present in real-world action sequences.

We begin by processing the input video $X$ to produce a tensor $\boldsymbol{V} \in \mathbb{R}^{L \times 3 \times H \times W}$, where $L$ represents a fixed sampling rate, and $H \times W$ is the resized dimension of each frame. To cater to actions occurring at varying temporal scales, we derive three types of subsequences, $\boldsymbol{V}^1$, $\boldsymbol{V}^4$, and $\boldsymbol{V}^8$, corresponding to scale sizes of 1, 4, and 8, respectively. This multi-scale approach ensures that our model is sensitive to both subtle and pronounced action repetitions, enabling it to accurately capture the intricate temporal dynamics present within the video sequences.

Utilizing the video Swin Transformer \cite{liu2022video}, we extract 3D features from each subsequence. To enrich these features with contextual information and capture long-range dependencies, a 3D convolutional network processes the extracted features, yielding enhanced representations $\widetilde{\boldsymbol{V}}^1$, $\widetilde{\boldsymbol{V}}^4$, and $\widetilde{\boldsymbol{V}}^8$ in $\mathbb{R}^{L \times 512}$. Additionally, a self-attention module computes the temporal self-similarity across all frames, producing matrices $\boldsymbol{S}^1$, $\boldsymbol{S}^4$, and $\boldsymbol{S}^8$ in $\mathbb{R}^{4 \times L \times L}$. These matrices capture the temporal relationships and dependencies within the video sequences at various scales.

To amalgamate the multi-scale information, we concatenate the matrices $\boldsymbol{S}^1$, $\boldsymbol{S}^4$, and $\boldsymbol{S}^8$, forming a comprehensive temporal self-similarity matrix $\boldsymbol{S}$, which encapsulates the temporal dynamics across all scales. This matrix $\boldsymbol{S}$ undergoes a 2D convolutional layer, followed by flattening and projection, resulting in a video feature tensor $\boldsymbol{E} \in \mathbb{R}^{L \times 512}$ that captures the rich spatial-temporal information present in the input video.

To further refine and integrate the extracted features, we employ a Transformer encoder with four attention heads and a 512-dimensional hidden layer. This encoder module allows for effective long-range dependency modeling and feature refinement, capturing the complex temporal relationships within the action sequences. The refined features from the Transformer encoder are then input into a three-layer fully connected network, which outputs the final action density map $\boldsymbol{p} \in \mathbb{R}^L$. This density map represents the model's prediction of action occurrences throughout the video, with each element $p_i$ corresponding to the predicted action density at the $i$-th frame.

To obtain the final predicted action count, we sum the elements of the density map:
\begin{equation}
    T = \sum_{i=1}^{L} p_i,
\end{equation}
where $T$ is the estimated total count of repetitive actions within the video. Following the process proposed by \cite{guo2011simple} and \cite{hu2022transrac}, we employ Gaussianization to construct the ground-truth density map $\boldsymbol{g} \in \mathbb{R}^L$, where $\overline{T} = \sum_{i=1}^{L} g_i$. This process involves applying a Gaussian function to the ground-truth action counts, with the mean $\mu$ positioned at the midpoint of each identified action cycle. This refined methodology for spatial-temporal encoding and prediction not only captures a broad spectrum of action characteristics across videos but also aligns the model's predictions closely with the nuanced temporal patterns of action repetitions.

Through the integration of multi-scale processing, advanced feature extraction techniques, temporal self-similarity analysis, and Transformer-based encoding, our approach sets a robust foundation for accurate and class-agnostic action counting. This comprehensive spatial-temporal encoder and prediction head architecture, combined with the novel loss formulation detailed in Section~\ref{subsec:irregular_priors}, enables our model to effectively handle the diverse and irregular nature of action repetitions present in real-world video sequences.

\subsection{Random Count Augmentation Strategy}
\label{subsec:rca}
The intrinsic variability and complexity of real-world videos pose a formidable challenge to the robustness and generalization capabilities of Video Action Counting (VAC) models. A key factor exacerbating this challenge is the scarcity of finely annotated datasets, which often fail to represent the wide spectrum of action sequences and count distributions encountered in practical scenarios. To mitigate this limitation and significantly enhance the adaptability of our IVAC-$\mathtt{P^2L}$ model across diverse VAC scenarios, we introduce an advanced data augmentation strategy termed Enhanced Random Count Augmentation (RCA). This strategy is meticulously designed to artificially augment the diversity and complexity of the training dataset, thereby substantially improving the model's performance on underrepresented action sequences and count distributions.

The operational principles of RCA is predicated on the strategic adjustment of ground-truth counts of action repetitions within training videos. This adjustment is guided by a calculated threshold, $\tau$, which epitomizes the average action count across the dataset $D$. The augmentation process unfolds through the following meticulously structured steps:
\begin{enumerate}
    \item Initially, for each video $X_i$ within the dataset, the ground-truth action count, $\overline{T_i}$, is ascertained.
    \item Subsequently, the average action count $\tau$ is calculated across all videos within the dataset $D$, serving as a benchmark for augmentation decisions.
    \item In cases where $\overline{T_i} \ge \tau$, a new count value is randomly sampled from the inclusive range $[1, \tau]$ to yield $\overline{T_i}^{new}$. This strategy aims to introduce variability by reducing the action count in videos with originally high counts. Conversely, for videos whose counts are already below $\tau$, the original count is retained to preserve the natural diversity of the dataset.
    \item To align the video $X_i$ with the newly generated count $\overline{T_i}^{new}$, the corresponding frames are either cropped or extended. This manipulation ensures that the modified video instance accurately reflects the adjusted action count, thereby enriching the dataset with a broader spectrum of action sequences.
\end{enumerate}

Mathematically, this augmentation logic is succinctly encapsulated as:
\begin{numcases}{\overline{T_i}^{new}=}
    \text{random sample}(1, \tau), & if $\overline{T_i} \ge \tau$; \label{eq:enhanced_rca_high}\\
    \overline{T_i}, & otherwise. \label{eq:enhanced_rca_low}
\end{numcases}
Typically, the introduction of RCA into our training regimen significantly enriches the dataset with a more extensive array of action counts, thereby exposing the model to a wider variety of action frequencies and sequences. This strategic exposure is instrumental in averting overfitting to specific count patterns, thereby significantly enhancing the model's predictive accuracy in videos characterized by either sparse or densely packed actions. Furthermore, the infusion of variability in action counts through RCA stimulates the development of a more nuanced understanding of action dynamics within the model. This, in turn, supports its generalization capabilities, empowering it to adeptly navigate unseen or novel VAC tasks.

The implementation of RCA necessitates a rational consideration of the dataset's intrinsic characteristics and the architectural nuances of the model. Required parameters such as the threshold $\tau$ and the probability of resampling must be meticulously optimized based on empirical performance metrics to ensure the efficacy of the augmentation. Moreover, it is crucial to ensure that the modified video instances retain coherent action sequences post-augmentation. This protection is essential for preserving the integrity and pedagogical value of the training data, thereby ensuring that the model learns from realistic and contextually sound video instances.

\section{Experimental Results and Analysis}

\subsection{Datasets}
\label{subsec:datasets}
In our evaluation of the IVAC-$\mathtt{P^2L}$ model for the video action counting task, we elected three widely recognized datasets that offer diverse challenges and cover a broad spectrum of real-world scenarios. Table~\ref{tab1} provides a summary of the key statistics of these datasets, which are RepCount \cite{hu2022transrac}, UCFRep \cite{zhang2020context}, and Countix \cite{dwibedi2020counting}. The selection criteria for these datasets were based on their varying degrees of complexity, annotation granularity, and application domains, ensuring a robust and thorough assessment of our proposed model's performance and versatility.

\textbf{Countix}: Introduced by Dwibedi et al. \cite{dwibedi2020counting}, the Countix dataset focuses exclusively on the labeling of repetitive action counts within videos. This dataset is derived from a wide array of YouTube videos, covering a vast range of activities. It is meticulously partitioned into training, validation, and testing sets comprising 4,588, 1,450, and 2,719 videos, respectively. The Countix dataset challenges the model with its diverse content and real-world variability, testing the model's ability to generalize across different settings without explicit temporal annotations for action cycles.

\textbf{UCFRep}: The UCFRep dataset, developed by Zhang et al. \cite{zhang2020context}, is an extension of the well-known UCF101 dataset \cite{soomro2012ucf101}. It consists of 526 videos, with a division of 421 videos for training and 105 for validation. Unlike Countix, UCFRep provides a controlled environment with videos derived from the UCF101 dataset, which is primarily focused on action recognition. The inclusion of UCFRep in our evaluation allows us to assess the model's effectiveness in extracting and counting repetitive actions from videos where the primary focus is not on repetition, thereby testing the adaptability of IVAC-$\mathtt{P^2L}$ to varied video contexts.

\textbf{RepCount}: The most recently introduced dataset, RepCount \cite{hu2022transrac}, offers fine-grained annotations that include the start and end moments of each action cycle within the videos. It contains a total of 1,451 videos, divided into two subsets: RepCount-A and Repcount-B. For our experiments, we utilize only the RepCount-A subset, as Repcount-B has not been made publicly available. RepCount-A includes 1,041 videos, which are further divided into training (758 videos), validation (131 videos), and testing sets (152 videos). The RepCount dataset poses a unique challenge by providing detailed temporal annotations for each repetitive action, facilitating a more nuanced evaluation of our model's capacity to precisely localize and count repetitive actions within a video.

\begin{table}[htbp]
    \caption{The statistics of RepCount-A, UCFRep and Countix datasets.}
    \centering
    \begin{tabular*}{\columnwidth}{@{\extracolsep{\fill}}lccc}
        \toprule
        {Dataset}                      & RepCount-A     & UCFRep          & Countix            \\
        \midrule                          
        Videos                         & 1041           & 526             & 8757               \\
        Avg Duration                   & 30.7           & 8.2             & 6.1                \\
        Min/Max Duration               & 4.0/88.0       & 2.1/33.8        & 0.2/10.0           \\
        Avg Cycles                     & 15.0           & 6.7             & 6.8                \\
        Min/Max Cycles                 & 1/141          & 3/54            & 2/73               \\              
        \bottomrule
    \end{tabular*}
\label{tab1}
\end{table}

These datasets collectively encompass a comprehensive range of video action counting scenarios, from simple repetitive motions to complex sequences with intricate temporal dynamics. Our selection of RepCount, UCFRep, and Countix enables a holistic evaluation of IVAC-$\mathtt{P^2L}$, testing its robustness, accuracy, and generalizability across different challenges and application domains in video action counting.

\subsection{Implementation Details}
\label{subsec:implementation_details}
For the training and evaluation of our IVAC-$\mathtt{P^2L}$ model, we meticulously crafted our experimental setup to ensure robustness and reproducibility. This section delineates the technical specifics of our implementation, including the optimization algorithm, data preprocessing, and augmentation techniques.

\subsubsection{Optimization and Training Settings}
Our model's optimization leverages the Adam optimizer, renowned for its effectiveness in handling sparse gradients and adaptive learning rates. We selected a batch size of 64 to balance between computational efficiency and the model's ability to generalize from the training data. Each input video is uniformly sampled to $L=64$ frames to standardize input dimensions, catering to both computational constraints and the necessity to capture sufficient temporal information. Subsequently, each frame is resized to $224 \times 224$ pixels, aligning with common practices in image and video processing tasks for neural networks. The learning rate and other hyperparameters of the Adam optimizer were set following empirical best practices and preliminary experiments to ensure stable convergence.

The loss function weights, $\alpha$, $\beta$, and $\gamma$, are crucial for balancing the contributions of different components of our model's learning objective. After extensive experimentation, these weights were uniformly set to 1, indicating equal importance across the loss components in the initial stages of training. This choice was guided by our goal to equally emphasize the relevance of cycle consistency, interval inconsistency, and regression accuracy in the early phases of model training.

\subsubsection{Data Augmentation}
Given the limited size of the RepCount-A training set, which comprises only 758 videos, we employed the Random Count Augmentation (RCA) strategy to enhance dataset diversity and model robustness. The RCA strategy is particularly designed to address the challenge posed by over-representation of certain action counts. For videos with a number of cycle segments exceeding the dataset's average value $\tau=15$, as detailed in Table~\ref{tab1}, we implemented a probabilistic resampling mechanism. With a probability of $0.5$, the action count for such videos is randomly resampled from a uniform distribution within the range $[1, \tau]$. This process not only introduces variation in action counts but also encourages the model to learn from a wider spectrum of video dynamics by either cropping or extending the video frames accordingly to reflect the new sampled count.

The implementation and training of our IVAC-$\mathtt{P^2L}$ model was conducted on the training and validation sets of the RepCount-A dataset. The evaluation was carried out on the testing set of RepCount-A, as well as the validation sets of UCFRep and Countix, to assess the model's performance across different data distributions and challenges inherent in the video action counting domain. Our experiments are designed to ensure that our IVAC-$\mathtt{P^2L}$ model not only learns effectively from the available data but also generalizes well across diverse video content. By adhering to these implementation details and utilizing strategic data augmentation, we aim to set a new benchmark in video action counting.

\subsection{Baselines}
\label{subsec:baselines}
To evaluate the performance of our IVAC-$\mathtt{P^2L}$ model, we selected diverse baseline methods that represent the current state-of-the-art in video action counting, as well as related fields such as action recognition and segmentation. This comparison not only benchmarks our model against established methods but also highlights the advancements made by IVAC-$\mathtt{P^2L}$. Below, we describe each baseline method and the rationale behind its selection.

\textbf{RepNet} \cite{dwibedi2020counting}: Developed by Dwibedi et al., RepNet stands as a seminal work in video action counting, employing a novel approach that predicts the cycle length for each frame individually and classifies each frame as either within a cycle or not. This method's ability to handle videos with varying repetition rates and irregular cycle durations makes it a pertinent benchmark for assessing the efficacy of our model in similar conditions.

\textbf{Zhang et al.} \cite{zhang2020context}: This method introduces a sophisticated technique for cycle count prediction by estimating the locations of identical frames across preceding and succeeding cycles. Its innovative use of contextual information to infer repetitive patterns provides a compelling comparison for evaluating the performance of our IVAC-$\mathtt{P^2L}$ model, especially in terms of temporal understanding and cycle identification.

\textbf{TransRAC} \cite{hu2022transrac}: As one of the most recent advances in the field, TransRAC adopts a comprehensive approach to count repetitive cycles by predicting a density map derived from multi-scale video features. This method's integration of spatial-temporal features for counting serves as an essential benchmark for our model, given our emphasis on leveraging irregular repetition priors for improved counting accuracy.

Additionally, we extend our comparison to include methods from closely related domains, specifically action recognition and segmentation. These methods are adapted to the video action counting task by modifying their output layers, thereby providing a broader context for evaluating the versatility and performance of our approach:

\textbf{X3D} \cite{feichtenhofer2020x3d}: A highly efficient and scalable version of the 3D ConvNet architecture that excels in video action recognition tasks. Its adaptability to different computational budgets makes it an interesting candidate for our comparison.

\textbf{TANet} \cite{liu2021tam}: Incorporates temporal aggregation to improve action recognition, highlighting the significance of effectively capturing temporal dynamics within video sequences.

\textbf{Video SwinT} \cite{liu2022video}: Applies the Swin Transformer to video understanding, showcasing the potential of transformer-based models in capturing complex spatial-temporal relationships.

\textbf{Huang et al.} \cite{huang2020improving}: Focuses on enhancing action segmentation through more effective feature learning, offering insights into the temporal segmentation in counting tasks.

By comparing these methods, we aim to demonstrate the efficacy of our approach in capturing and counting repetitive actions across diverse video datasets.

\subsection{Evaluation Metrics}
\label{subsec:evaluation_metrics}
To ensure a thorough and meaningful evaluation of the performance across different video action counting methods, including our proposed IVAC-$\mathtt{P^2L}$ model, we employ two widely accepted evaluation metrics that have been established in prior works \cite{dwibedi2020counting, zhang2020context, hu2022transrac}. These metrics, Mean Absolute Error (MAE) and Off-By-One Accuracy (OBO), offer complementary insights into the accuracy and reliability of action count predictions.

\textbf{Mean Absolute Error (MAE)}: The MAE metric quantifies the average magnitude of errors in the predicted counts, without considering their direction. It is defined as the normalized absolute difference between the predicted count ($T_i$) and the ground-truth count ($\overline{T_i}$) across all videos in the dataset. The normalization by the ground-truth count provides a relative error measure that is more interpretable across videos with varying numbers of actions. Mathematically, MAE is expressed as follows:
\begin{equation}
    \centering
    \label{eq12}
    \mathtt{MAE} = \frac{1}{K} \sum_{i=1}^{K} \frac{| T_i - \overline{T_i} |}{\overline{T_i}},
\end{equation}
where $K$ represents the total number of videos in the dataset. A lower MAE value indicates a higher accuracy of the model in predicting the exact counts of repetitive actions, making it a crucial metric for evaluating the precision of counting algorithms.

\textbf{Off-By-One Accuracy (OBO)}: In addition to precise count accuracy, it is also informative to assess the model's performance in terms of its ability to produce counts that are close to the ground truth. The OBO metric addresses this by measuring the proportion of predictions that are within one count of the actual value, effectively capturing the model's tolerance for minor inaccuracies. The OBO is defined as:
\begin{equation}
    \centering
    \label{eq13}
    \mathtt{OBO} = \frac{1}{K} \sum_{i=1}^{K} \mathbbm{1}\left[ |T_i - \overline{T_i}| \leq 1 \right],
\end{equation}
where $\mathbbm{1}[\cdot]$ is the indicator function, returning 1 when the condition inside is true, and 0 otherwise. This metric provides a lenient assessment of count predictions, emphasizing the practical usability of the model in applications where absolute precision is less critical than general accuracy.

Together, MAE and OBO offer a comprehensive evaluation of model performance, with MAE emphasizing precision and OBO assessing the model's predictive reliability within a practical margin of error. By adopting these metrics, we align our evaluation with established standards in the field, facilitating direct comparisons with existing and future methods in video action counting.

\begin{table}[htbp]
\caption{Performance comparison of different approaches on the RepCount-A dataset. (*) denotes the results obtained by re-running the officially released code with the proposed augmentation strategy on the RepCount-A dataset.}
\centering
    \begin{tabular*}{\columnwidth}{@{\extracolsep{\fill}}lcc}
    \toprule
    \multirow{2}{*}{Algorithms}                      &\multicolumn{2}{c}{RepCount-A}                 \\
                                                     \cmidrule(r){2-3}                               
                                                     &\multicolumn{1}{c}{MAE ($\downarrow$)}         &\multicolumn{1}{c}{OBO ($\%$, $\uparrow$)}        \\ 
    \midrule
    X3D \cite{feichtenhofer2020x3d}                  & 0.9105              & 10.59                   \\
    TANet \cite{liu2021tam}                          & 0.6624              &  9.93                   \\    
    Video SwinT \cite{liu2022video}                  & 0.5756              & 13.24                   \\ 
    Huang et al. \cite{huang2020improving}           & 0.5267              & 15.89                   \\ 
    Zhang et al. \cite{zhang2020context}             & 0.8786              & 15.54                   \\ 
    RepNet \cite{dwibedi2020counting}                & 0.9950              &  1.34                   \\
    \midrule
    TransRAC \cite{hu2022transrac} (*)               & 0.4158              & 25.83                   \\
	\textbf{IVAC-$\mathbf{P^2L}$}                    & \textbf{0.4022}     & \textbf{34.44}          \\
    \bottomrule
	\end{tabular*}
\label{tab2}
\end{table}

\begin{table}[htbp]
\caption{Performance comparison of different approaches on the UCFRep and Countix datasets when trained on the augmented RepCount-A dataset (RepNet is trained on RepCount-A). (*) denotes the results obtained by re-running officially released codes with the data augmentation on RepCount-A.}
\centering
    \begin{tabular*}{\columnwidth}{@{\extracolsep{\fill}}lcccc}
    \toprule
    \multirow{2}{*}{Algorithms}                      &\multicolumn{2}{c}{UCFRep}                     &\multicolumn{2}{c}{Countix}       \\
                                                     \cmidrule(r){2-3}                               \cmidrule(l){4-5}
                                                     &\multicolumn{1}{c}{MAE ($\downarrow$)}         &\multicolumn{1}{c}{OBO ($\%$, $\uparrow$)}    &\multicolumn{1}{c}{MAE ($\downarrow$)}         &\multicolumn{1}{c}{OBO ($\%$, $\uparrow$)}                 \\ 
    \midrule
    RepNet \cite{dwibedi2020counting}                & 0.9985              & 0.90                    & -                   & -                    \\
    TransRAC \cite{hu2022transrac} (*)               & 0.5961              & 32.00                   & 0.5742              & 38.48                \\
    \textbf{IVAC-$\mathbf{P^2L}$}                    & \textbf{0.5028}     & \textbf{42.00}          & \textbf{0.5071}     & \textbf{43.09}       \\
    \bottomrule
	\end{tabular*}
\label{tab3}
\end{table}

\subsection{Quantitative Results}
\label{subsec:quantitative_results}
Our extensive evaluation encompasses three datasets, each presenting unique challenges to video action counting methods. This section details the quantitative performance of our proposed IVAC-$\mathtt{P^2L}$ model in comparison to a range of baseline methods, highlighting its effectiveness across diverse video contexts. Detailed results are as follows.

\textbf{Performance on RepCount-A}: 
The RepCount-A dataset, with its fine-grained annotations, serves as a critical benchmark for assessing the precision of action counting models. As illustrated in Table~\ref{tab2}, our IVAC-$\mathtt{P^2L}$ model demonstrates superior performance, outperforming the state-of-the-art TransRAC \cite{hu2022transrac} method by a margin of 0.0136 in MAE and 8.61\% in OBO. Notably, against the second-best method by Huang et al. \cite{huang2020improving}, our model achieves significant improvements of 0.1245 in MAE and 18.55\% in OBO, underscoring its robustness and accuracy in identifying and counting repetitive actions within videos.

\textbf{Performance on UCFRep}: 
The UCFRep dataset, derived from the action recognition-centric UCF101, poses a distinct challenge due to its diverse action representations and less frequent cycle annotations. Despite these challenges, as shown in Table~\ref{tab3}, IVAC-$\mathtt{P^2L}$ markedly surpasses the TransRAC model, improving by 0.0933 in MAE and 10.0\% in OBO. This enhancement emphasizes the adaptability of our approach to datasets where repetitive actions are embedded within complex video sequences.

\textbf{Performance on Countix}: 
The Countix, with its extensive and varied video content, tests the generalization capability of counting models. Our approach, as reported in Table~\ref{tab3}, achieves significant performance boosts over TransRAC, with a 0.0671 improvement in MAE and a 4.61\% increase in OBO. The substantial size of the Countix validation set, nearly ten times larger than RepCount-A's testing set, further validates the scalability and generalizability of the IVAC-$\mathtt{P^2L}$ model across different video domains.

These results demonstrate the effectiveness of the IVAC-$\mathtt{P^2L}$ model in addressing the intricacies of video action counting. By outperforming established methods across three datasets, our approach demonstrates not only the utility of leveraging irregular repetition priors but also its potential to advance the state-of-the-art in video action counting research. The quantitative findings underscore our model's capability to accurately count actions in videos with varying levels of complexity and annotation detail, positioning it as a versatile tool for both research and practical applications in the field.

\begin{table}[h]
\caption{The ablation of number of phases in the cycle segment on the RepCount-A dataset.}
\centering
    \begin{tabular*}{\columnwidth}{@{\extracolsep{\fill}}ccc}
        \toprule
        \multirow{2}{*}{Phases}&\multicolumn{2}{c}{RepCount-A}  \\
        \cmidrule(r){2-3}
        &\multicolumn{1}{c}{MAE ($\downarrow$)}     &\multicolumn{1}{c}{OBO ($\%$, $\uparrow$)}            \\ 
        \midrule
	   3                            & 0.4270               & 32.45             \\
	   2                            & 0.4399               & 32.45             \\
	   \textbf{1}                   & \textbf{0.4022}      & \textbf{34.44}    \\
        \bottomrule
	\end{tabular*}
\label{table_4}
\end{table}

\subsection{Ablation Studies}
In this section, we conduct ablation studies for the number of phases in achieving the inter-cycle consistency, the effectiveness of the Pull and Push losses, the performance comparison of different variants of the pull-push loss, the effectiveness of the proposed augmentation strategy on RepCount-A and the effect of different sampling rates.

\subsubsection{Number of Phases in Cycle}
Concerning achieving the inter-cycle semantic consistency, we further demonstrate the performance of different granularity-level alignments in the pull loss. 
Specifically, we divide each cycle segment into multiple phases, representing different processes in executing an action. 
For example, in the three-phase case, three phases of a cycle segment correspond to the start, middle and end processes of an action. 

The reference embedding of the $j-$th phase in the $h-$th cycle segment is defined as:
\begin{equation}
    \centering
    \label{eq14}
    \mathcal R_{h,j} = \frac{1}{|\mathcal C_{h,j}|} \sum_{i \in \mathcal C_{h,j}} \boldsymbol E_{i,:},
\end{equation}
where $\mathcal C_{h,j}$ denotes the set of frames belonging to the $j-$th phase of the $h-$th cycle segment in the video $X$. 
Then, the reference embedding of the $j-$th phase of all cycle segments in the video $X$ is computed as follows:
\begin{equation}
    \centering
    \label{eq15}
    \mathcal R^{j} = \frac{1}{C} \sum_{h=1}^{C} \mathcal R_{h,j}.
\end{equation}
The pull loss is updated as:
\begin{equation}
    \centering
    \label{eq16}
    \mathcal L_{Pull} = \frac{1}{C} \sum_{h=1}^{C} \left(\sum_{j=1}^{M} \left(1 - {\cos} \left(\mathcal R_{h,j}, \mathcal R^{j} \right) \right) \right),
\end{equation}
where $M$ is the number of phases in a cycle segment.

We report the results of two and three phases in Table~\ref{table_4}.
In the two-phase case, one can observe that the performance drops by $0.0377$ on MAE and $1.99\%$ on OBO. 
In the three-phase case, we can see that the MAE metric decreases by $0.0248$ and the OBO metric drops by $1.99\%$. 
The above results show that partitioning an entire cycle segment into different phases leads to inferior performance. The potential explanation is that the partition of a cycle segment causes the semantic separation of an action, which further results in unsatisfactory performance in achieving the inter-cycle embedding consistency.

\begin{table}[htbp]
\caption{The ablation of the pull loss and the push loss on the RepCount-A dataset.}
\centering
    \begin{tabular*}{\columnwidth}{@{\extracolsep{\fill}}cccc}
        \toprule
        \multirow{2}{*}{Pull}     &\multirow{2}{*}{Push}   &\multicolumn{2}{c}{RepCount-A}  \\
        \cmidrule(r){3-4}
            &   &\multicolumn{1}{c}{MAE ($\downarrow$)}     &\multicolumn{1}{c}{OBO ($\%$, $\uparrow$)}            \\ 
        \midrule
                        &                      & 0.4158            & 25.83             \\
	                  & \checkmark           & 0.4285            & 32.45             \\
	  \checkmark      &                      & 0.4253            & 31.13             \\
	  \checkmark      & \checkmark           & \textbf{0.4022}   & \textbf{34.44}    \\
        \bottomrule
	\end{tabular*}
\label{table_5}
\end{table}

\subsubsection{Effectiveness of Pull-Push Loss}
\label{subsubsec:effectiveness_pull_push}
To validate the contribution of our proposed pull-push loss mechanism within the IVAC-$\mathtt{P^2L}$ framework, we conducted an ablation study by individually deactivating the pull and push components. The performance impact of these modifications was quantitatively assessed on the RepCount-A dataset, with the findings reported in Table~\ref{table_5}.

Table~\ref{table_5} reveals that omitting the pull loss results in a decline of 0.0263 in MAE and 1.99\% in OBO, while excluding the push loss leads to a reduction of 0.0231 in MAE and 3.31\% in OBO. These observations underscore the critical role of both loss components in our model's architecture. Notably, the more pronounced impact of removing the push loss highlights the significance of distinguishing between cycle and interval segments to achieve precise action counting.

\begin{table}[htbp]
\caption{The Ablation of Variants of the Pull-Push Loss on the RepCount-A Dataset.}
\centering
    \begin{tabular*}{\columnwidth}{@{\extracolsep{\fill}}lcc}
        \toprule
        \multirow{2}{*}{Loss}                             &\multicolumn{2}{c}{RepCount-A}                  \\
        \cmidrule(r){2-3}
        &\multicolumn{1}{c}{MAE ($\downarrow$)}           &\multicolumn{1}{c}{OBO ($\%$, $\uparrow$)}      \\ 
        \midrule
        Contrastive Loss           & 0.4803             & 31.13             \\
        Triplet Loss               & 0.4693             & 33.11             \\
        \textbf{Pull-Push Loss}    & \textbf{0.4022}    & \textbf{34.44}    \\
        \bottomrule
	\end{tabular*}
\label{table_6}
\end{table}

\begin{figure*}[htbp]
\centering
    \includegraphics[width=0.95\textwidth]{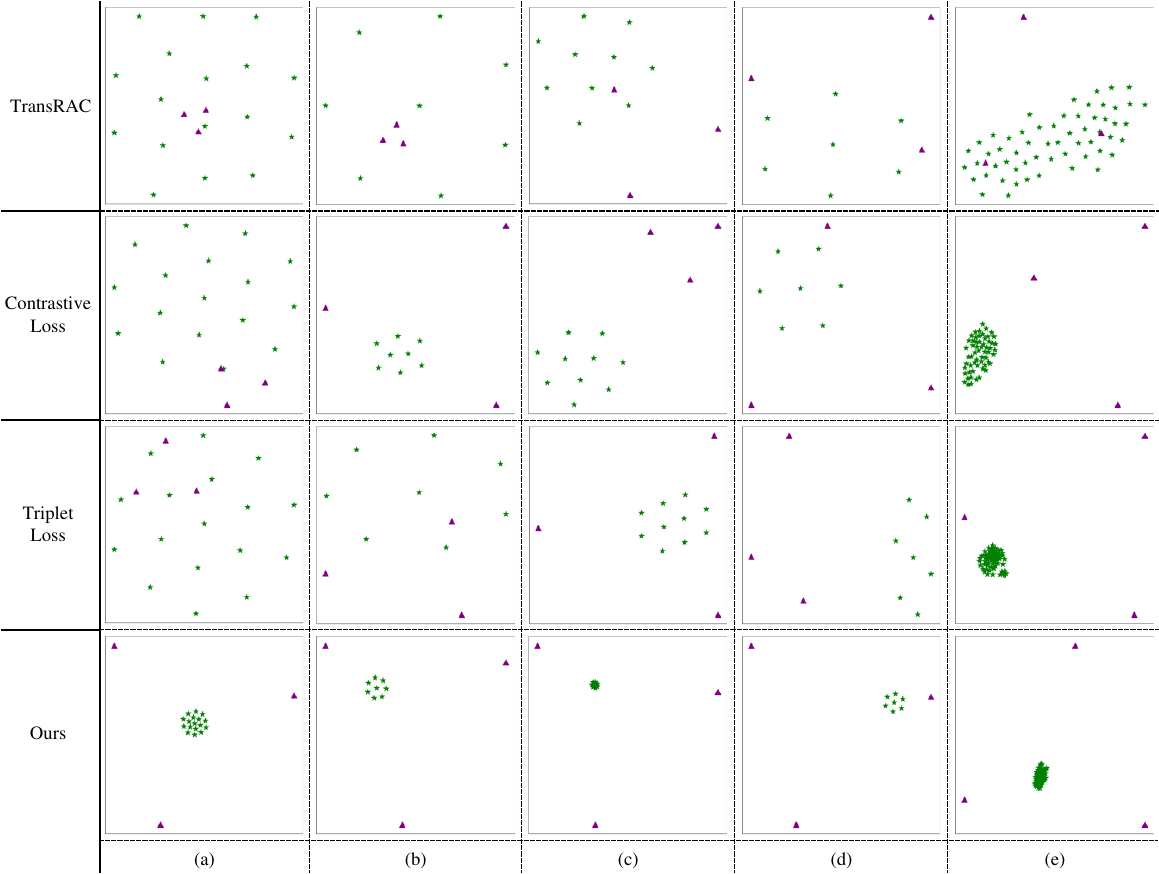} 
\caption{Comparative t-SNE visualization of feature embeddings across various video action counting methods on the RepCount-A dataset. Each column showcases embeddings from a single video, illustrating the distribution of cycle and interval segments as perceived by different models. The green stars denote the aggregated embeddings for cycle segments, symbolizing repetitive actions within the video, whereas the purple triangles indicate the embeddings for interval segments, representing non-repetitive or distinct actions. This visualization underscores the efficacy of our approach in achieving clear separation and clustering of cycle and interval segments in the embedding space, thereby highlighting the advantages of our method in distinguishing between repetitive and non-repetitive video segments with enhanced precision.}
\label{fig4}
\end{figure*}

\subsubsection{Variants of the Pull-Push Loss}
Regarding achieving the inter-cycle consistency and the cycle-interval inconsistency, we also make some variants for the pull-push loss. Specifically, we replace the pull-push loss with the contrastive loss \cite{hadsell2006dimensionality} and the triplet loss \cite{schroff2015facenet}, and report the comparison results in Table~\ref{table_6}.

\textbf{Contrastive Loss.}
The contrastive loss is widely used in unsupervised representation learning, aiming at closing the representation distance of positive pairs while expanding the representation distance of negative pairs.
Motivated by InfoNCE \cite{oord2018representation}, the contrastive loss adapted to the video action counting task can be expressed as follows:
\begin{equation}
    \centering
    \label{eq17}
    \mathcal L_{Contras} = \frac{1}{C} \sum_{h=1}^{C} - \log \left( \frac{e^{\cos \left( \mathcal R_{h}, \mathcal R \right)} / \tau}{\sum_{j=1}^{C+N} e^{\cos \left( \mathcal R_{h}, \hat{\mathcal R}_j \right)} / \tau} \right),
\end{equation}
where $\{\hat{\mathcal R}_j\} = \{\mathcal R_h\} \cup \{\widetilde{\mathcal R}_k\}$, $\tau$ is the temperature rate and set as $0.07$ in experiments. 
As shown in Table~\ref{table_6}, it can be observed that the MAE and OBO metrics decrease by large margins of $0.0781$ and $3.31\%$, respectively, compared with the pull-push loss.
One plausible explanation for the poor performance of the contrastive loss is that, typically, the number of interval segments in a video is significantly smaller than the number of cycle segments. Consequently, this can lead to a substantial reduction in the number of negative pairs compared to the number of positive pairs, which will greatly reduce the prediction performance.

\textbf{Triplet Loss.}
The triplet loss \cite{schroff2015facenet} aims to make the distance between the features of the same category as close as possible, and the distance between the features of different categories as far as possible, and for the positive and negative examples of the same category, let the distance between them be greater than the margin.
Inspired by the graphical loss defined in \cite{zhang2019graphical}, the triplet loss in video action counting can be written as follows:
\begin{equation}
    \centering
    \label{eq18}
    \Phi_h = \min \limits_{c \in [1,C], c \ne h} {\cos}(\mathcal R_h, \mathcal R_c) - \max \limits_{k \in [1,N]} {\cos} (\mathcal R_{h}, \widetilde {\mathcal R}_{k}),
\end{equation}
\begin{equation}
    \centering
    \label{eq19}
    \mathcal L_{Triplet} = \max(0, \lambda - \Phi_h),
\end{equation}
where we aim to maximize the similarity of the lowest-scoring positive pair and minimize the similarity of the highest-scoring negative pair. 
The parameter $\lambda$ represents the margin threshold, and in our experiments, we have chosen $\lambda=2$ to achieve the best performance.
As we can see from Table~\ref{table_6}, the MAE metric drops by $0.0671$ and the OBO metric drops by $1.33\%$ compared with the pull-push loss.
A plausible explanation for the aforementioned results is that for sample pairs with large inter-class differences, the triplet loss becomes $0$ during training. This circumstance can result in the results being trapped in a local optimum, leading to a decrease in the overall performance of the trained model.

The aforementioned results demonstrate that, as compared to other variant losses, the pull-push loss achieves the optimal outcome in capturing the irregular repetitive priors in the video action counting task.

\begin{table}[htbp]
\caption{The ablation of data augmentation on the RepCount-A dataset.}
\centering
    \begin{tabular*}{\columnwidth}{@{\extracolsep{\fill}}lcc}
        \toprule
                            &\multicolumn{1}{c}{MAE ($\downarrow$)}      &\multicolumn{1}{c}{OBO ($\%$, $\uparrow$)}   \\ 
        \midrule
        TransRAC (w/o aug)                 & 0.4474                      & 23.18                                       \\
        TransRAC                           & 0.4158                      & 25.83                                       \\
        IVAC-$\mathtt{P^2L}$ (w/o aug)     & \textbf{0.3976}             & 32.45                                       \\
	  \textbf{IVAC-$\mathbf{P^2L}$}      & 0.4022                      & \textbf{34.44}                              \\
        \bottomrule
	\end{tabular*}
\label{table_7}
\end{table}

\subsubsection{Effectiveness of Dataset Augmentation}
We verify the effectiveness of our designed data augmentation strategy on the RepCount-A dataset, and report the results in Table~\ref{table_7}. We denote these methods without the dataset augmentation as ``w/o aug''.

Regarding the TransRAC method, incorporating the dataset augmentation will bring $0.0316$ and $2.65\%$ improvements on MAE and OBO, respectively. 
Compared with IVAC-$\mathtt{P^2L}$ (w/o aug) on the RepCount-A dataset, the data augmentation of IVAC-$\mathtt{P^2L}$ brings 0.0153 improvements in the average performance, taking into account both MAE and OBO metrics.
These results effectively verify the effectiveness of our designed augmentation strategy on the RepCount-A dataset.

\begin{table}[htbp]
\caption{The ablation of sampling rate on the RepCount-A dataset.}
\centering
    \begin{tabular*}{\columnwidth}{@{\extracolsep{\fill}}lcc}
        \toprule
        Sampling Rate                                     &\multicolumn{1}{c}{MAE ($\downarrow$)}    &\multicolumn{1}{c}{OBO ($\%$, $\uparrow$)}   \\ 
        \midrule
	  IVAC-$\mathtt{P^2L}$ ($L=64$)                   & 0.4022                                   & \textbf{34.44}                   \\
        \textbf{IVAC-$\mathbf{P^2L}$ ($L=128$)}         & \textbf{0.3799}                          & 32.45                            \\
        \bottomrule
	\end{tabular*}
\label{table_8}
\end{table}

\subsubsection{Impact of Sampling Rate }
\label{subsubsec:impact_sampling_rate}
Recognizing the pivotal role of sampling rate in the feature encoding process, we examined the performance of IVAC-$\mathtt{P^2L}$ under different frame sampling rates, specifically $L=64$ and $L=128$. This investigation aimed to ascertain the optimal sampling rate that maximizes the model's efficacy in capturing detailed video features essential for accurate action counting. The comparative analysis, presented in Table~\ref{table_8}, indicates that a sampling rate of $L=128$ significantly enhances the model's performance, achieving a 0.0223 improvement in MAE, albeit with a slight 1.99\% decrease in OBO. This finding elucidates the importance of a higher sampling rate in extracting more granular and informative video features, thereby facilitating more precise action recognition and counting. The slight trade-off observed in OBO suggests a nuanced balance between capturing detailed features and maintaining a broad accuracy margin in predictions, which warrants further investigation in future research endeavors.

\begin{figure}[htbp]
\centering
    \includegraphics[width=0.98\columnwidth]{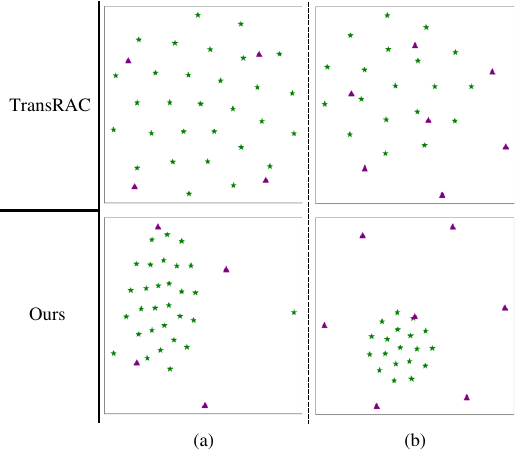} 
\caption{t-SNE visualizations highlighting failure cases in video action counting on the RepCount-A dataset. Each column visualizes feature embeddings from the same video, detailing instances where our model did not achieve optimal segmentation between cycle and interval segments. The green stars and purple triangles represent the reference embeddings of cycle and interval segments, respectively. These visualizations elucidate the challenges faced in distinguishing between repetitive and non-repetitive actions under certain conditions, providing insights into areas for further improvement and refinement of our approach.}
\label{fig5}
\end{figure}

\subsection{Visualization and Qualitative Analysis}
We utilized the t-SNE algorithm to visualize reference embeddings of cycle and interval segments from the RepCount-A dataset, comparing IVAC-$\mathbf{P^2L}$ with TransRAC, contrastive, and triplet loss methods as shown in Fig.~\ref{fig4}. Additionally, Fig.~\ref{fig5} presents some instances where our method did not perform optimally.

Observations from Fig.~\ref{fig4} reveal that IVAC-$\mathbf{P^2L}$ distinctively enhances the separation between cycle and interval segments, surpassing the performance of TransRAC, contrastive, and triplet loss techniques. Specifically, in Fig.~\ref{fig4}(a), while cycle and interval segments appear nearly indistinguishable under the other methods, IVAC-$\mathbf{P^2L}$ successfully clusters cycle segments, clearly differentiating them from interval segments. This trend persists across scenarios with a limited number of cycles (Fig.~\ref{fig4}(b)-(d)), where our approach consistently outperforms the alternatives. Notably, even in complex scenarios (Fig.~\ref{fig4}(e)), where contrastive and triplet losses show some efficacy, IVAC-$\mathbf{P^2L}$ achieves superior segmentation.

Failure cases highlighted in Fig.~\ref{fig5} suggest limitations of the pull-push loss under certain conditions. For instance, Fig.~\ref{fig5}(a) shows inadequate separation between cycle and interval segments compared to TransRAC. This is attributed to significant variances within cycle segments and minimal differences between cycle and interval segments. Similarly, in Fig.~\ref{fig5}(b), despite effective clustering of cycle segments, the push loss fails to maintain a clear distinction from interval segments. These findings demonstrate the efficacy of the pull-push loss in promoting inter-cycle consistency and cycle-interval inconsistency, showcasing its advantages over contrastive and triplet losses in video action counting tasks.

\section{Conclusion}
This study presents a framework, IVAC-$\mathbf{P^2L}$, leveraging irregular repetition priors to advance video action counting accuracy. By innovatively addressing the challenges of variable-length cycle segments and interruptions, we establish two foundational principles: Inter-cycle Consistency and Cycle-interval Inconsistency. These principles guide the creation of our consistency and inconsistency modules and the pioneering pull-push loss mechanism, $\mathbf{P^2L}$. The pull loss promotes uniform representation across cycle segments for similar actions, whereas the push loss distinguishes these from interval segments, identifying distinct actions. Extensive evaluation on the RepCount-A dataset confirmed our method's superiority, notably outperforming TransRAC and showing remarkable generalizability across the Countix and UCFRep datasets without specific tuning. These results not only validate our framework's effectiveness and adaptability but also signify a major stride forward in video action counting research, especially in handling irregular repetition patterns and varied action dynamics.

\section*{Acknowledgments}
This research was partially supported by the National Natural Science Foundation of China (Grant No. T2341003). The contributions made by Zhi-Qi Cheng were partially funded by the financial assistance award 60NANB17D156 from the U.S. Department of Commerce, National Institute of Standards and Technology (NIST).

\bibliographystyle{IEEEtran}
\bibliography{ref}

\begin{IEEEbiography}[{\includegraphics[width=1in,height=1.25in,clip,keepaspectratio]{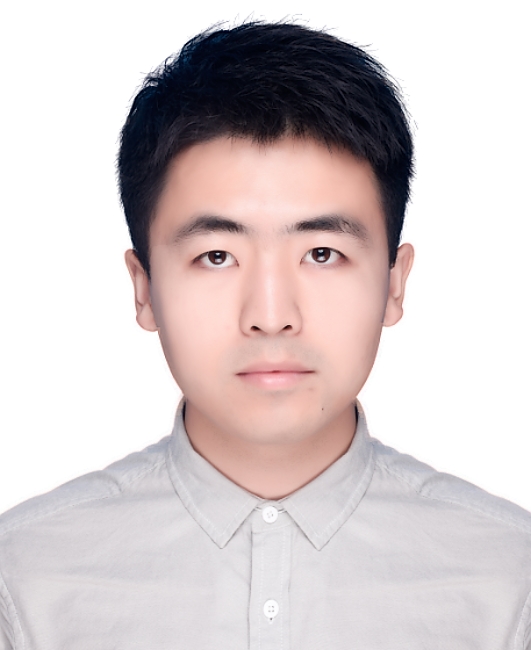}}]{Hang Wang}
received his bachelor's degree in the School of Computer Science and Technology from Xi'an Jiaotong University in 2017, and later earned his Ph.D. from the School of Cyber Science and Engineering at the same university in 2023.
Currently, he is an assistant professor at the School of Automation Science and Engineering, Xi'an Jiaotong University. He is also pursuing the second Ph.D. degree with the Department of Computing at The Hong Kong Polytechnic University. His research interests include machine learning, computer vision and image processing.
\end{IEEEbiography}

\begin{IEEEbiography}[{\includegraphics[width=1in,height=1.25in,clip,keepaspectratio]{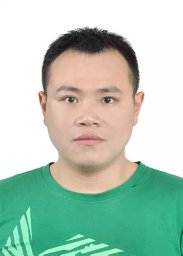}}]{Zhi-Qi Cheng} 
received his B.S. in Computer Science from Southwest Jiaotong University in 2014 and completed his Ph.D. in 2019. During his doctoral studies, he was a joint Ph.D. student at the City University of Hong Kong (2016-2017) and later at Carnegie Mellon University (2017-2019). He also gained valuable experience through internships at Alibaba DAMO Academy (2016), Google Brain (2018), and Microsoft Research (2019). From 2019 to 2022, he served as a postdoctoral research associate at the School of Computer Science of Carnegie Mellon University (CMU). He is currently a Project Scientist at the Language Technologies Institute (LTI), part of the School of Computer Science at CMU. He significantly contributed to several key projects, including DARPA's AIDA, KAIROS, IARPA's DIVA, and NIST's PSIAP. His research has had a substantial impact, such as being utilized in the Washington Post's coverage of the Capitol riots, for which he was awarded the Pulitzer Prize for Public Service. Additionally, he has been honored with the Intel Ph.D. Fellowship and IBM Outstanding Student Scholarship.
\end{IEEEbiography}

\begin{IEEEbiography}[{\includegraphics[width=1in,height=1.25in,clip,keepaspectratio]{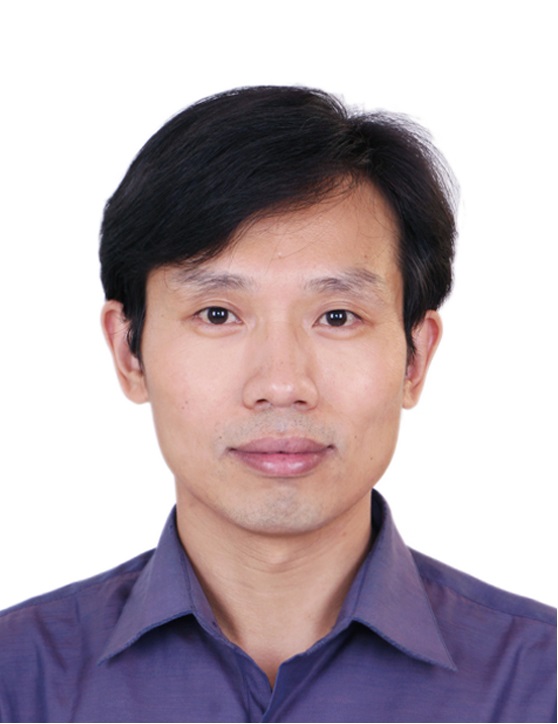}}]{Youtian Du}
received the Bachelor degree from Xi'an Jiaotong University, and Ph.D. degree from Tsinghua University, in 2002 and 2008, respectively. He was a visiting scholar at the Department of Electrical and Computer Engineering, the University of Massachusetts, Amherst in 2016-2017. He is currently a Professor at the Faculty of Electronic and Information Engineering, Xi'an Jiaotong University. His current research interests mainly include cross-media analysis and reasoning, knowledge extracting and representation, and machine learning.
\end{IEEEbiography}

\begin{IEEEbiography}[{\includegraphics[width=1in,height=1.25in,clip,keepaspectratio]{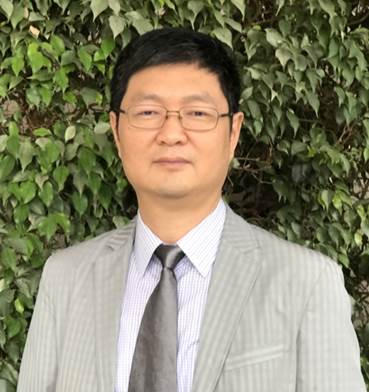}}]{Lei Zhang}
(M’04, SM’14, F’18) received his B.Sc. degree in 1995 from Shenyang Institute of Aeronautical Engineering, Shenyang, P.R. China, and M.Sc. and Ph.D. degrees in Control Theory and Engineering from Northwestern Polytechnical University, Xi’an, P.R. China, in 1998 and 2001, respectively. From 2001 to 2002, he was a research associate in the Department of Computing, The Hong Kong Polytechnic University. From January 2003 to January 2006 he worked as a Postdoctoral Fellow in the Department of Electrical and Computer Engineering, McMaster University, Canada. In 2006, he joined the Department of Computing, The Hong Kong Polytechnic University, as an Assistant Professor. Since July 2017, he has been a Chair Professor in the same department. His research interests include Computer Vision, Image and Video Analysis, Pattern Recognition, and Biometrics, etc. Prof. Zhang has published more than 200 papers in those areas. As of 2022, his publications have been cited more than 80,000 times in literature. Prof. Zhang is a Senior Associate Editor of IEEE Trans. on Image Processing, and is/was an Associate Editor of IEEE Trans. on Pattern Analysis and Machine Intelligence, SIAM Journal of Imaging Sciences, IEEE Trans. on CSVT, and Image and Vision Computing, etc. He is a ``Clarivate Analytics Highly Cited Researcher" from 2015 to 2022. More information can be found in his homepage \url{https://www4.comp.polyu.edu.hk/~cslzhang/}.
\end{IEEEbiography}
\end{document}